\documentclass[10pt,twocolumn,letterpaper]{article}

\usepackage{cvpr}              % To produce the CAMERA-READY version
\usepackage[dvipsnames]{xcolor}

\definecolor{cvprblue}{rgb}{0.21,0.49,0.74}
\usepackage[pagebackref,breaklinks,colorlinks,citecolor=cvprblue]{hyperref}
\usepackage{amssymb}
\usepackage{bbm}
\usepackage{adjustbox}
\usepackage{multirow}
\usepackage{bm}

\def\paperID{} % *** Enter the Paper ID here
\def\confName{}
\def\confYear{2025}

\title{
DreamActor-H1: High-Fidelity Human-Product Demonstration Video Generation via Motion-designed Diffusion Transformers
}

\author{
  $\text{Lizhen Wang}^1$, $\text{Zhurong Xia}^1$\textsuperscript{*}, $\text{Tianshu Hu}^1$, 
  $\text{Pengrui Wang}^1$, $\text{Pengfei Wei}^1$, \\
  $\text{Zerong Zheng}^1$, $\text{Ming Zhou}^1$, $\text{Yuan Zhang}^1$, $\text{Mingyuan Gao}^1$
  \\
  ${}^1$ ByteDance Intelligent Creation
  \\
  \tt\small \{wanglizhen.2024, zhaogang.666, tianshu.hu, wangpengrui.chj, pengfei.wei, \\
  \tt\small zerong, zhouming.9527\}@bytedance.com
  \tt\small \{zhang.yuan09,gaomingyuan001\}@gmail.com
}
\begin{document}

\twocolumn[{%
\renewcommand\twocolumn[1][]{#1}%
\maketitle
\begin{center}
    \centering
    \captionsetup{type=figure}
    \includegraphics[width=0.99\textwidth]{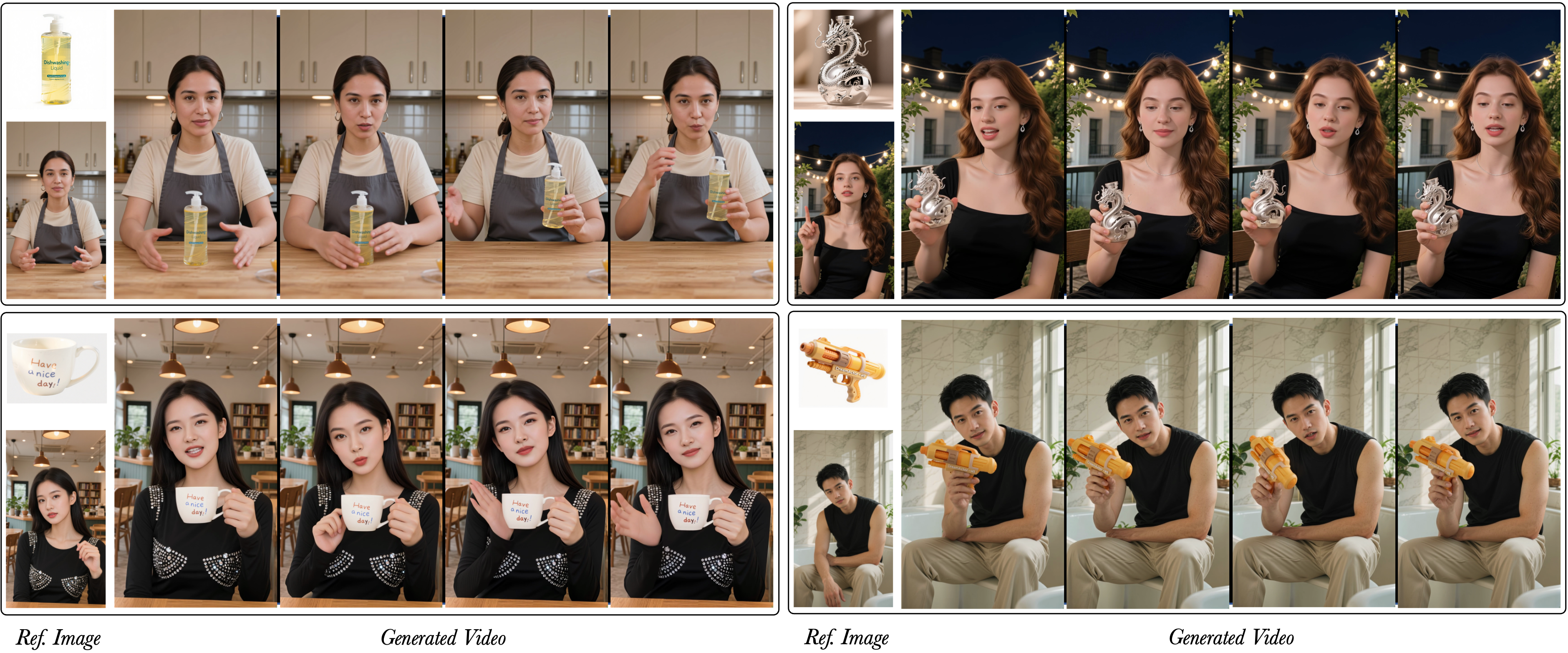}
    \captionof{figure}{DreamActor-H1 can generate high-fidelity and photo-realistic human-product demonstration videos from human and product reference images.}
    \label{fig:teaser}
\end{center}%
}]

\let\thefootnote\relax\footnotetext{* Corresponding author \& project leader}

\begin{abstract}
In e-commerce and digital marketing, generating high-fidelity human-product demonstration videos is important for effective product presentation. However, most existing frameworks either fail to preserve the identities of both humans and products or lack an understanding of human-product spatial relationships, leading to unrealistic representations and unnatural interactions.
To address these challenges, we propose a Diffusion Transformer (DiT)-based framework. Our method simultaneously preserves human identities and product-specific details, such as logos and textures, by injecting paired human-product reference information and utilizing an additional masked cross-attention mechanism. We employ a 3D body mesh template and product bounding boxes to provide precise motion guidance, enabling intuitive alignment of hand gestures with product placements. Additionally, structured text encoding is used to incorporate category-level semantics, enhancing 3D consistency during small rotational changes across frames. Trained on a hybrid dataset with extensive data augmentation strategies, our approach outperforms state-of-the-art techniques in maintaining the identity integrity of both humans and products and generating realistic demonstration motions. Project page: https://lizhenwangt.github.io/DreamActor-H1/.
\end{abstract}

\section{Introduction}
\label{sec:intro}

With the advancement of image-to-video generation technologies, it has become possible to produce human-product demonstration videos from just a product image and a person image. In the era of e-commerce, intuitively conveying product features through natural human motions would significantly enhance product marketing. However, despite the technological progress and growing demand, existing methods still struggle to simultaneously preserve both human identity and product details while ensuring correct product-human interactions with natural motions, which limits their practical applications.

% Current methods for video generation often focus on pose-driven human animation or object manipulation in isolation. While pose-transfer techniques achieve impressive human motion realism, they typically overlook the seamless integration of specific product appearances and semantic information, leading to poor preservation of object identities (e.g., logos, textures, or shapes). Other works that attempt to replace handheld objects in videos either rely on limited 2D image editing, which fails to maintain 3D consistency, or adopt generic multi-object generation frameworks that sacrifice product fidelity for motion complexity. Moreover, these approaches lack user-controllable mechanisms for precise alignment of human gestures and product placements, making them unsuitable for scenarios requiring fine-grained customization.

Despite advancements in pose-driven animation and object manipulation, existing human-product interaction video generation methods fall short of practical needs. First, traditional hand-object interaction (HOI)-focused approaches like HOI-Swap~\cite{xue2024hoiswap} are limited to partial hand-object scenarios, neglecting full-body dynamics. Recent HOI-oriented video generation methods such as AnchorCrafter~\cite{xu2024anchorcrafter} and Re-HOLD~\cite{fan2025ReHOLD} incorporate object and motion information into diffusion models, facilitating object or human replacement in videos. However, they still suffer from low-quality outputs and reliance on predefined motions, which limits them to size-constrained product replacement.
Second, pose-guided frameworks (e.g., MimicMotion~\cite{zhang2024mimicmotion}) do not focus on modeling human-product relationships. When generating videos from human-product images, they often produce severe object deformation as handheld items follow body movements. Finally, while text-to-image diffusion-based methods like InteractDiffusion~\cite{Hoe_2024_CVPR} and PersonaHOI~\cite{hu2025personahoi} generate HOI images from text prompts, they suffer from vague object details due to semantic prompt limitations. Multi-subject customization methods like Phantom~\cite{liu2025phantom} have shown promising potential in generating multi-subject customized videos, but they also rely on mapping subject images to anchored text prompts, introducing semantic ambiguity that fundamentally limits their capacity to model human-product spatial relationships.

To tackle these challenges, we introduce DreamActor-H1, a novel Diffusion Transformer (DiT)~\cite{peebles2023scalable}-based framework that generates high-quality human-product demonstration videos from paired human and product images. By integrating masked cross-attention to fuse appearance features from both inputs, our method preserves fine-grained details such as human identities, product logos, textures, and contours with high fidelity. For motion realism, we employ a 3D body mesh template alongside product bounding boxes to guide hand gestures and object placements, ensuring natural alignment of human interactions with products of varying sizes and shapes. Additionally, structured text encoding injects category-level semantics, enabling the model to learn shared visual features across product classes and enhance the material visual quality and 3D consistency during small rotations. Trained on a large-scale hybrid dataset with multi-class augmentation, DreamActor-H1 outperforms state-of-the-art methods in preserving human-product identity integrity and generating physically plausible demonstration motions, making it suitable for personalized e-commerce advertising and interactive media. To conclude, Our key contributions include:
\begin{itemize}
\item We propose a DiT-based framework to generate high-quality human-product demonstration videos from paired images, preserving good human and object identities via our appearance guidance module.
\item We use 3D body templates and product bounding boxes as motion guidance with an automatic matching algorithm, enhancing user-friendliness and practicality.
\item Our method employs structured text to enhance material visual quality and maintain 3D consistency during subtle product rotations.
\end{itemize}

% \item We propose a DiT-based framework that generates high-quality human-product demonstration videos from paired human-product images, which demonstrates robust human and object identity preservation with our appearance guidance module.
% \item We utilize a 3D-body template and product bounding boxes as motion guidance, accompanied by an automatic matching algorithm, which is more user-friendly and practical.
% \item Our method leverages product-related semantic text to enhance the material visual quality and 3D consistency by injecting structured text.
    
\section{Related Work}
\label{sec:related}

\subsection{Human-Object Interaction}
The generation of humans interacting with objects has been a fundamental research goal.
While extensive methods~\cite{xu2025intermimic, huang_etal_cvpr25, Hassan:CVPR:2021, hassan2023synthesizing, zhang2020place, zhang2020generating, diller2024cghoi, zhang2024graspxl, li2023object, peng2023hoi, xu2023interdiff, ghosh2022imos, liu2024easyhoi, ye2023vhoi} have focused on predicting human motions during interactions with objects based on 3D representations, recent advances~\cite{hu2022hand, ye2023affordance, zhang2024hoidiffusion, HOIGen, xue2024hoiswap, zhao2025tasterobadvancingvideogeneration, Hoe_2024_CVPR, hu2025personahoi, chen2024virtualmodel, xu2024anchorcrafter, fan2025ReHOLD} have begun to address the challenge of human-object interactions (HOI) in image and video generation tasks.
HOGAN~\cite{hu2022hand} first explores hand-object interaction image generation via a split-and-combine approach considering inter-occlusion, while HOI-Swap~\cite{xue2024hoiswap} proposes a two-stage hand-centric video editing framework for object swapping with HOI awareness but limited to single-hand grasping.
To generate full-body human-object interactions, InteractDiffusion~\cite{Hoe_2024_CVPR} provides interaction controllability while PersonaHOI~\cite{hu2025personahoi} enables facial personalization, both building upon pre-trained text-to-image diffusion models.
VirtualModel~\cite{chen2024virtualmodel} generates HOI images by leveraging input objects and human poses.
To facilitate human-object interactions in video generation, HOIGen-1M~\cite{HOIGen} introduces a large-scale dataset for HOI generation with over one million high-quality videos.
AnchorCrafter~\cite{xu2024anchorcrafter} integrates HOI into pose-guided human video generation models, while ignoring the compatibility of pose sequences with objects of various sizes and shapes.
Re-HOLD~\cite{fan2025ReHOLD} uses specialized hand-object layout representation to disentangle hand modeling and object adaptation for motion sequences, but its focus on video object replacement limits generative freedom.
% Leveraging specialized layout representation for hands and objects, Re-HOLD~\cite{fan2025ReHOLD} effectively disentangles hand modeling and object adaptation to various motion sequences. However, it focuses primarily on video object replacement, which leads to a limited ability to generate videos with sufficient degrees of freedom.
% HOGAN~\cite{hu2022hand} first explores hand-object interactions image generation with the consideration of the inter-occlusion and generates the target image in a split-and-combine manner.
% HOI-Swap~\cite{xue2024hoiswap} proposes a two-staged hand-centric video editing framework to achieve object swapping with HOI awareness, but only supports single-hand grasping operations

\subsection{Pose-Guided Video Generation}
Given a human image and pose sequence, pose-guided human video generation~\cite{karras2023dreampose, xu2024magicanimate, chang2023magicpose, wang2024disco, ma2024follow, wang2024unianimate, zhu2024champ, huang2024make, hu2024animate, zhang2024mimicmotion, tu2024stableanimator, li2024dispose, peng2024controlnext, guan2024talk, gan2025humandit, luo2025dreamactor, hu2025animate, men2024mimo} aims to synthesize human animations following the pose sequence.
To improve pose guidance accuracy and flexibility, researchers have explored various pose representations such as 2D skeleton map~\cite{chang2023magicpose, wang2024disco, ma2024follow, wang2024unianimate}, 3D human parametric model~\cite{zhu2024champ, huang2024make}, and dense correspondences~\cite{karras2023dreampose, xu2024magicanimate}.
In addition, AnimateAnyone~\cite{hu2024animate} proposes ReferenceNet to effectively preserve intricate visual consistency.
Building upon a video diffusion model~\cite{blattmann2023stable}, MimicMotion~\cite{zhang2024mimicmotion} designs a confidence-aware mechanism to address model instability from keypoint uncertainty while StableAnimator~\cite{tu2024stableanimator} develops specialized modules for identity consistency.
DisPose~\cite{li2024dispose} and ControlNeXT~\cite{peng2024controlnext} introduce plug-and-play modules to seamlessly integrate into existing models, maintaining training efficiency.
TALK-Act~\cite{guan2024talk} enhances textural awareness with explicit motion guidance to tackle the difficulty of synthesizing stable hand movements.
Building upon diffusion transformer~\cite{peebles2023scalable}, HumanDiT~\cite{gan2025humandit} facilitates learning for long-form video generation.
DreamActor-M1~\cite{luo2025dreamactor} proposes the carefully designed hybrid guidance to achieve fine-grained holistic controllability, multi-scale adaptability, and long-term temporal coherence.
Notably absent from prior work is explicit modeling of human-object interactions, creating a substantial gap for real-world physical scenarios. 
Apart from human pose, MIMO~\cite{men2024mimo} and Animate Anyone 2~\cite{hu2025animate} take into account environmental representations as additional input conditions, aiming to effectively characterize the relationship between character and environment.
However, they introduce visual artifacts when dealing with complex hand-object interactions that occupy a relatively small pixel region, leading to suboptimal performance.

\begin{figure*}
  \centering
  \includegraphics[width=\linewidth]{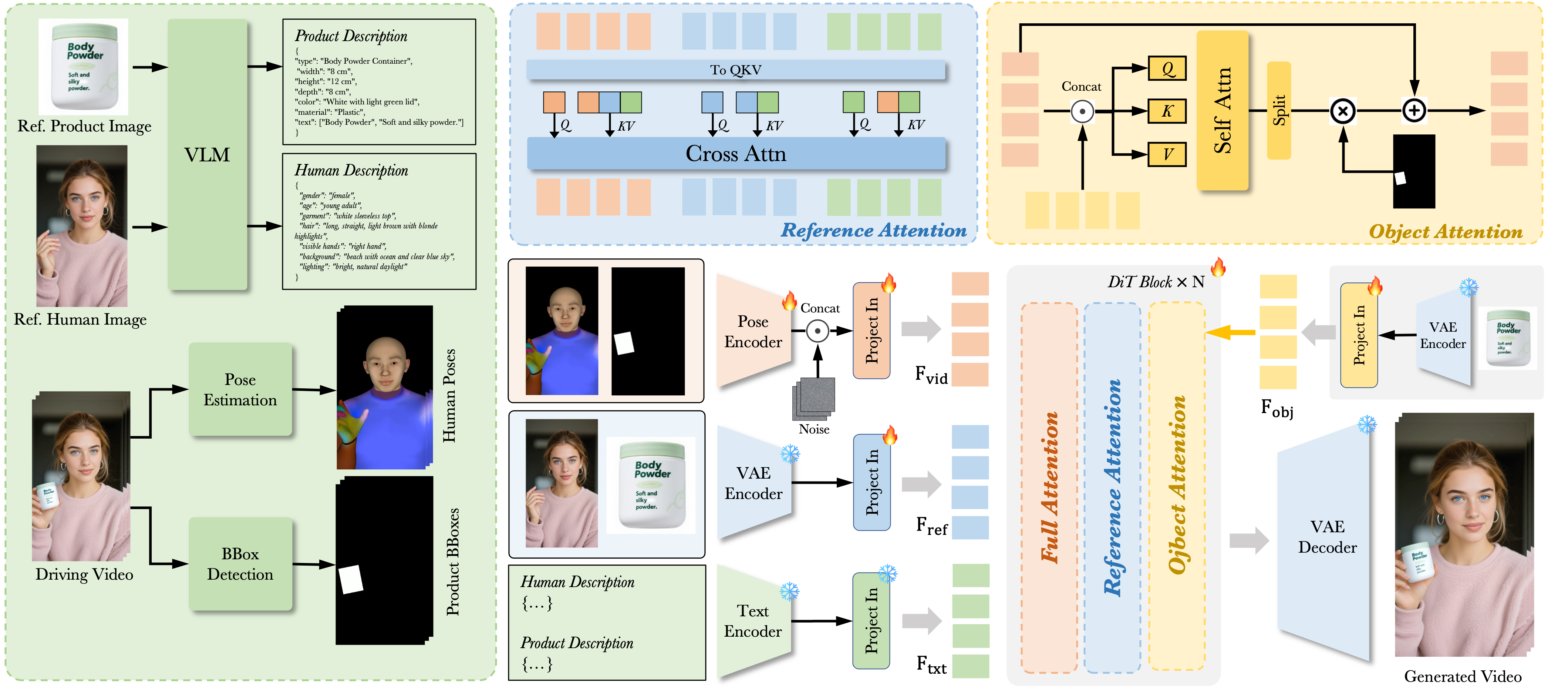} %0.95
  % \vspace{-0.3 cm}
  \caption{The pipeline of DreamActor-H1 leverages a DiT architecture, starting with dataset preparation where a VLM describes product and human images, followed by pose estimation and bounding box detection on training videos. During training, human poses and product boxes integrate with video noise for motion guidance, while a VAE encodes input images for appearance guidance; human-product descriptions are fed into the model via a text encoder. The model incorporates full attention, reference attention, and object attention (with product latents as inputs), with the reference and object attention mechanisms detailed at the top of the figure.}
  % \vspace{-0.3 cm}
  % \Description{}
  \label{fig:pipeline}
\end{figure*}

\subsection{Customized Video Generation}
Customized video generation~\cite{he2024id, yuan2024identity, zhong2025concat, wei2024dreamvideo, jiang2023videobooth, kim2025subject, chen2025multi, liang2025movie, huang2025videomage, huang2025conceptmaster, liu2025phantom, vace, deng2025cinema, hu2025hunyuancustom} aims to produce videos featuring specific subjects and maintaining identity consistency across different contexts.
Some previous works~\cite{he2024id, yuan2024identity, zhong2025concat} focus on maintaining facial identity. For instance, ID-Animator~\cite{he2024id} introduces a face adapter to encode the ID-relevant embeddings from learnable facial latent queries and ConsisID~\cite{yuan2024identity} designs a tuning-free pipeline equipped with frequency-aware identity-control scheme.
Besides, DreamVideo~\cite{wei2024dreamvideo} learns to customize both subject identity and motion upon finetuning on a few images and videos.
VideoBooth~\cite{jiang2023videobooth} proposes a feed-forward framework by embedding image prompts in a coarse-to-fine manner.
Through decoupling the subject-specific learning from temporal dynamics, Kim et al.~\cite{kim2025subject} propose a recipe to train the customized video generation model without annotated videos.
Recent advances such as VideoAlchemist~\cite{chen2025multi}, MovieWeaver~\cite{liang2025movie}, and Phantom~\cite{liu2025phantom} have demonstrated potential in multi-subject video generation. However, these methods mainly rely on mapping subject images to text prompts, introducing semantic ambiguity that limits inter-subject relationship modeling. To tackle this, CINEMA~\cite{deng2025cinema} and HunyuanCustom~\cite{hu2025hunyuancustom} use multimodal large language models (MLLMs) to enhance interactive integration of subject images and text, improving cross-modal alignment.
However, no prior methods have been specifically designed for human-product demonstrations, resulting in significant challenges in preserving fine-grained product details such as text and logos.
% Recent advances like VideoAlchemist~\cite{chen2025multi}, MovieWeaver~\cite{liang2025movie}, ConceptMaster~\cite{huang2025conceptmaster}, Phantom~\cite{liu2025phantom}, VideoMage~\cite{huang2025videomage} and VACE~\cite{vace} have shown promising potential in generating multi-subject customized videos. 
% Nevertheless, these methods primarily depend on mapping subject images to anchored text prompts, introducing semantic ambiguity that fundamentally limits their capacity to model inter-subject relationships.
% To address this challenge, methods including CINEMA~\cite{deng2025cinema} and HunyuanCustom~\cite{hu2025hunyuancustom} leverage multimodal large language model (MLLM) to enhance the interactive integration of subject images and text entities, thereby improving cross-modal alignment.
\section{Method}
\label{sec:method}

\subsection{Overview}

% DreamActor-H1 can generate a high-fidelity human-product demonstration video from a human reference image and a product reference image. As shown in Fig.~\ref{fig:pipeline}, our framework is based on a DiT architecture, specifically, Seaweed-7B~\cite{seawead2025seaweed} a foundational model for video generation with approximately 7 billion (7B) parameters. During the dataset preparation stage, we first use a VLM to describe the product and human images and then we utilize pose estimation and bounding box detection for the training product-human demonstration video. During the training stage, we concatenate the human pose and product box with the input video noise as the motion guidance. We then encode the input human and product images with a variational autoencoder (VAE) as appearance guidance. The human and product description will be treated as supplementary information to improve the material visual quality and 3D consistency during small rotational changes across frames. For the DiT model, we employ stacks of Full Attention, Reference Attention, and Object Attention. In this setup, Object Attention takes the product latent as additional input information. As shown in Fig.~\ref{fig:inference}, during the inference stage, we apply automatic pose template selection depending on the human and product information. Finally, as a whole, our method can address the challenges of identity preservation, motion realism, and spatial relationship modeling and generate a high quality human-product demonstration video with a human and a product image input.

\begin{figure}
  \centering
  \includegraphics[width=\linewidth]{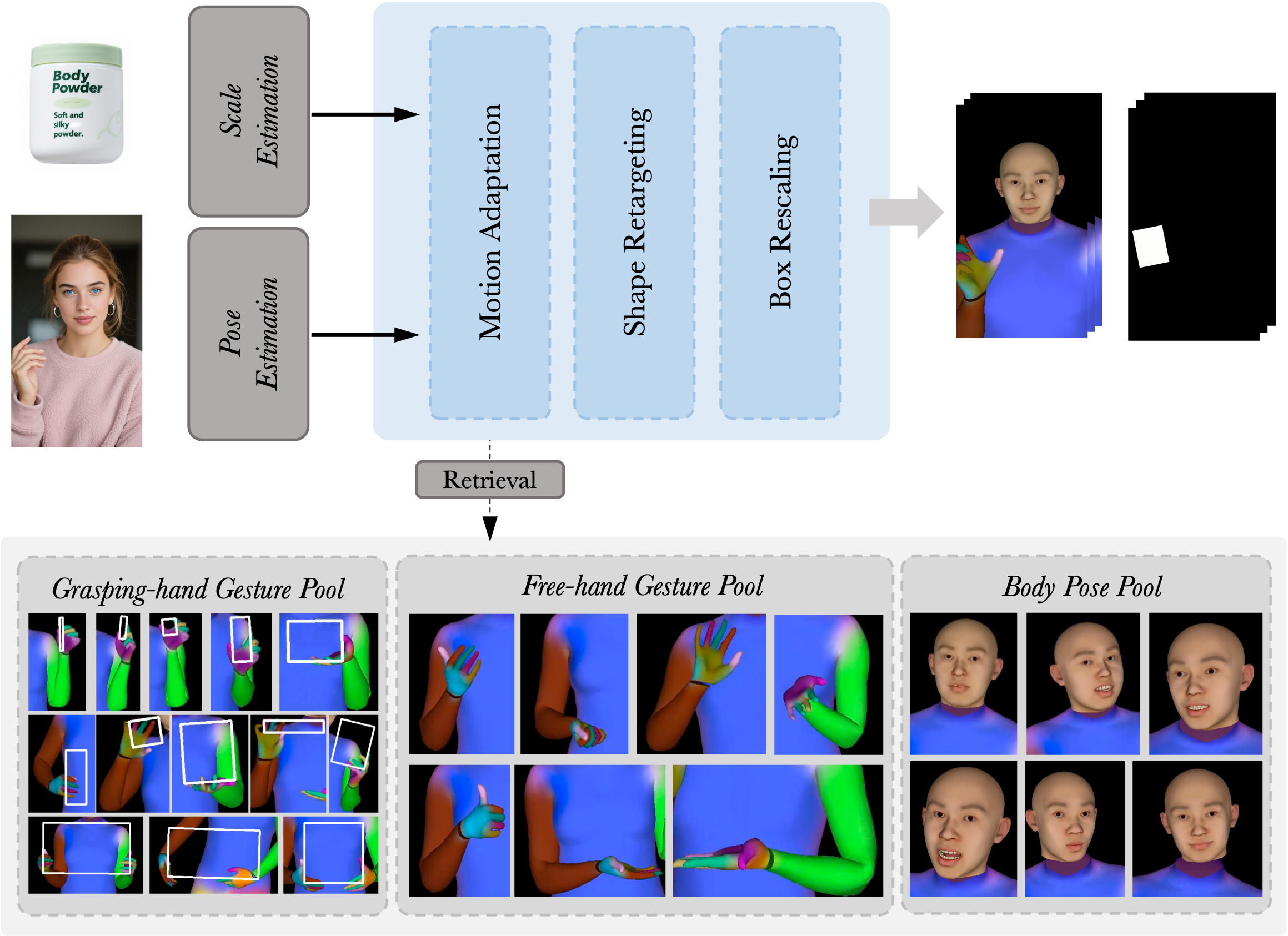} %0.97
  % \vspace{-0.2 cm}
  \caption{During inference, our framework retrieves optimal motion templates from pre-defined pools and adapts object box scaling via joint analysis of reference human/product images, enabling pose-coherent animations.}
  % \vspace{-0.2 cm}
  % \Description{}
  \label{fig:inference}
\end{figure}

DreamActor-H1 can generate high-fidelity human-product demonstration videos from a human reference image and a product reference image. As depicted in Fig.~\ref{fig:pipeline}, our framework is built upon a DiT architecture, specifically leveraging Seaweed-7B~\cite{seawead2025seaweed}, a foundational model for video generation with around 7 billion (7B) parameters.
In the dataset preparation phase, we use a Vision-Language Model (VLM) to describe the product and human images. Subsequently, pose estimation and bounding box detection are applied to the training product-human demonstration video.
During the training stage, we combine the human pose and product bounding box with the input video noise to serve as motion guidance, as described in Sec.~\ref{sec:motion}. Additionally, we encode the input human and product images using a variational autoencoder (VAE)~\cite{kingma2013auto} to serve as appearance guidance. The descriptions of the human and product are utilized as supplementary information, enhancing the material visual quality and 3D consistency during small rotational changes across frames, as explained in Sec.~\ref{sec:motion}.
Regarding the DiT model, we implement stacks of full attention, reference attention, and object attention, as described in Sec.~\ref{sec:app}. Notably, object attention incorporates the product latent as an extra input.
As shown in Fig.~\ref{fig:inference}, during the inference stage, we implement automatic pose template selection based on human and product information. Overall, our approach can overcome the challenges of identity preservation, motion realism, and spatial relationship modeling, and produce high-quality human-product demonstration videos given a human and a product image as inputs. 

\subsection{Human-Product Appearance Guidance}
\label{sec:app}

In video generation, we need to first inject product and human appearance information while maintaining consistency. Unlike image-to-video tasks and pose-based human video driving tasks, our approach involves integrating product image information into a human reference image. And in practical applications, maintaining the consistency of the product appearance is of great importance.

% In the initial injection stage, we refer to ~\cite{lin2025omnihuman, luo2025dreamactor} and use a reference attention for appearance injection, which can avoid employing a copy of the DiT as the ReferenceNet~\cite{hu2024animate} and inject the reference feature into the DiT with mostly the same parameters and a little additional parameters. By concatenating and combining with self-attention, the network extracts appearance information from the input reference image. However, in MMDiT ~\cite{esser2024scaling}, we found that as the network deepens, some detailed information of the product will be lost during the process of gradual self-update (such as the text on the product, or some small geometric structures of the product). Therefore, we additionally use a masked object attention, which injects the object latents obtained by encoding the product image through VAE. By concatenating and combining with self-attention, we inject the product information in the form of a residual, which can improve the consistency of product details in the generated video.

In the initial injection stage, we adopt reference attention for appearance injection, following~\cite{lin2025omnihuman, luo2025dreamactor}. This approach avoids the need for a standalone ReferenceNet~\cite{hu2024animate} by directly integrating reference features into the DiT with minimal additional parameters (mostly shared with the base model). By concatenating reference features with self-attention mechanisms, the network extracts appearance information from input reference images. However, we found that as the network deepens, some detailed information of the product (e.g. text labels, micro-geometric structures) will be lost during the process of gradual self-update in MMDiT~\cite{esser2024scaling}. Therefore, we additionally use a masked object attention, which injects the object latents obtained by encoding the product image through VAE. By concatenating and combining with self-attention, we inject the product information in the form of a residual, which can improve the consistency of product details in the generated video.

Specifically, in the full attention, the input ${F_{\text{vid}} \in \mathbb{R}^{(t \times h \times w) \times c}}$, ${F_{\text{ref}} \in \mathbb{R}^{(1 \times h \times w) \times c}}$, and ${F_{\text{txt}} \in \mathbb{R}^{l \times c}}$ are concatenated along the first dimension (which includes the flattened temporal dimension $t$, spatial dimensions $h \times w$, or the compressed token count $l$ of text) and then processed through full self-attention to produce outputs $F'_{\text{vid}}$, $F'_{\text{ref}}$, and $F'_{\text{txt}}$ of the same dimensions. 
In the reference attention, the inputs are reshaped as ${F_{\text{vid}} \in \mathbb{R}^{t \times (h \times w) \times c}}$, ${F_{\text{ref}} \in \mathbb{R}^{1 \times (h \times w) \times c}}$, and ${F_{\text{txt}} \in \mathbb{R}^{1 \times l \times c}}$, followed by separate attention operations along the second dimension (spatial dimensions $h, w$ per frame) to generate corresponding outputs. During the attention computation for ${F'_{\text{vid}}}$, queries ($Q$) are derived from ${F_{\text{vid}}}$, while keys and values ($KV$) are sourced from $F_{\text{vid}}$, $F_{\text{ref}}$, and $F_{\text{txt}}$. For ${F'_{\text{ref}}}$, $Q$ comes from ${F_{\text{ref}}}$ and $KV$ from $F_{\text{ref}}$ and $F_{\text{txt}}$. For ${F'_{\text{txt}}}$, $Q$ is from ${F_{\text{txt}}}$ and $KV$ from $F_{\text{vid}}$ and $F_{\text{txt}}$. This design ensures ${F_{\text{vid}}}$ integrates information from both ${F_{\text{ref}}}$ and ${F_{\text{txt}}}$, while ${F_{\text{ref}}}$ focuses solely on text correlations.
In the object attention, the input ${F_{\text{obj}} \in \mathbb{R}^{1 \times (h \times w) \times c}}$, a fixed VAE-encoded feature that does not self-update, is concatenated with ${F_{\text{vid}} \in \mathbb{R}^{t \times (h \times w) \times c}}$ along the second dimension and processed via self-attention. To selectively refine only product-containing regions in ${F_{\text{vid}}}$, the resulting features are scaled by a product region mask and added back to ${F_{\text{vid}}}$ as a residual update. As demonstrated in Sec.~\ref{subsec:ablation}, this attention design can significantly enhance the consistency of product appearance preservation.

\subsection{Human-Product Motion Guidance}
\label{sec:motion}

On one hand, substantial research has been conducted on human-object interaction (HOI), yet most efforts focus on 3D domains, typically requiring 3D models of objects. This poses significant limitations for practical applications, as the diversity of products makes it challenging to obtain realistic 3D models through image-to-3D techniques. On the other hand, common challenges include interactions that fail to align with text prompt instructions and notable inconsistencies in object sizes, which decrease the realism of visual outputs. To address these challenges, we implement a motion guidance method that combines 3D human templates with object bounding boxes to direct video generation. This strategy allows for more precise control over object positioning and scale adjustment.

In our pose estimation, we leverage the 3D body template SMPL-X~\cite{SMPL-X:2019} and face model FaceVerse~\cite{wang2022faceverse} to capture human movements, with 3D parameter fitting guided by frameworks including 4D-Humans~\cite{goel2023humans}, HaMeR~\cite{pavlakos2024hamer}, and FaceVerse~\cite{wang2022faceverse}. Note that we have manually colored SMPL-X mesh to enhance the visual differences between different regions and assist the network learning. For product localization, we employ a two-stage approach: initial bounding box detection using GroundingDINO~\cite{liu2024grounding} followed by refined segmentation via SAM2~\cite{ravi2024sam}. The final object bounding box is derived as the minimum rotated rectangle from the segmentation masks, enabling more accurate product size and representation of z-axis rotations compared to boxes from GroundingDINO. Our pose encoder, inspired by~\cite{hu2024animate}, uses a lightweight CNN architecture. During the training stage, our motion guidance is concatenated with input noise along the channel dimension, providing explicit spatial cues for video generation. The entire pose encoding module is trained end-to-end to optimize motion consistency.

During the inference stage, as shown in Fig.~\ref{fig:inference}, we design an automated logic for selecting and adjusting motion templates to generate the final video from a given single human and product image. To this end, we have constructed a motion template pool from the dataset, which includes movements such as two-hand and head motions, small-scale body movements during speaking, and object-related actions like single-hand grasping, lifting, picking up, raising, two-hand holding, and picking up products from tables, along with the corresponding changes in product bounding boxes, expandable directions of the bounding boxes, and fixed dimensions (e.g., fixing the width when the expansion direction is ``up'' or fixing the height for ``left and right''), where another free dimension can be adjusted according to the aspect ratio of the input product. The motion templates cover products of various sizes from 1 to 40 centimeters and some common demonstration actions. In practical applications, we first obtain the shape, translation, rotation information of the target human image and the product size information (predicted by VLM or manually assigned), then match the most suitable hand motion template for holding the object based on the product size and body orientation, and also match a suitable template for facial expressions and minor movements of the head and trunk according to the body orientation. If the object-holding hand only involves a single hand, we will match a demonstration action template for the other hand. We retain the human's position, shape, rotation and other information for retargeting according to the motion template, and the product bounding box will be readjusted to the corresponding position and size following the retargeted object-holding hand. If it involves two-hand actions, the product bounding box size needs to be readjusted according to the distance between the two hands. Finally, we also adjust the final product bounding box according to the freely expandable direction of the product marked in the motion template and the aspect ratio of the input product. After rendering, the motion guidance video for inference can be obtained.

\subsection{Product Semantics Guidance}
\label{sec:text}

After the above operations, we have successfully injected product image information into human images to generate demonstration videos with good product preservation. However, there are still issues in representing minor product rotations and material properties (such as reflections from transparent materials, plastics, or metals), and text on products may occasionally fail to be preserved. We speculate that this arises because the network lacks common-sense knowledge about products of the same category or material, making it difficult to proactively determine product categories and materials from a single input. 

Therefore, we used a VLM to generate text captions for product categories, sizes, colors, materials, and text, as well as annotations for human image attributes like human information, environment, and lighting. Here, we utilize Seed1.5-VL~\cite{seed2025seed1_5vl} as our annotation tool. Specifically, to highlight key attributes, we represent input text in a dictionary-like format that contains only essential keyword information, as shown in Fig.~\ref{fig:pipeline}. During training and inference, these texts are converted into text embeddings by the text encoder in Seaweed model to assist video generation. This allows the network to better learn common-sense information corresponding to text from same-category or same-material products in the dataset, thereby improving representation of material properties. Notably, this input also enhances stability during small product rotations, likely leveraging geometric patterns from same-category products in the training set.

% To enhance 3D consistency across frames and generalize to unseen product categories, we introduce structured text encoding:

% Category-Level Semantics: Product categories (e.g., “electronics,” “utensils”) are encoded into text embeddings using a pre-trained language model (e.g., CLIP~\cite{radford2021learning}).
% Semantic-Visual Fusion: The text embeddings are concatenated with visual features in the DiT’s cross-attention layers, enabling the model to learn category-specific interaction patterns.
% Rotational Consistency: During frame generation, the semantic embeddings help maintain stable object representations (e.g., a smartphone’s screen orientation) across small rotations, reducing deformation artifacts.

\subsection{Dataset and Training Details}

To enhance the network's capabilities in human-product identity preservation, generating dynamic motion changes, and maintaining 3D consistency of products, we adopt a hybrid dataset for model training. Our core self-collected data focuses on training motion and appearance guidance modules, while supplementary datasets address tasks like product rotation, background naturalness, and identity preservation. Specifically, we collected 15,000 pairs of product-human images and corresponding demonstration videos (approximately 80 hours), featuring over 5,000 structurally simple everyday products with actions covering standing/sitting and tabletop pickup poses. This primary dataset contributes 80\% of the training samples. Additionally, we captured rotation videos for each product to assist 3D structure prediction from frontal images. To enrich human identity diversity and dynamic backgrounds (e.g., ocean waves), we collected 50 hours of e-commerce live-stream data and incorporated 100 hours of generalized categories from the Seaweed dataset. These supplementary datasets, lacking paired images, contribute 20\% of samples as simple image-to-video tasks.

For broader applicability across different human formats (full-body, half-body, selfies), we applied random crop augmentation and color augmentation to adjust lighting, contrast and saturation. We further utilized IC-Light~\cite{zhang2025scaling} to simulate diverse lighting conditions, ensuring robustness to varying product photography environments.
During training, we employ Flow Matching~\cite{lipman2022flow} as the training objective with region-specific weighting for faces, hands and products.
% \begin{equation}
% \mathcal{L} = \mathbb{E}_{t \sim \mathcal{U}(0,1)} \left[ w(\mathbf{x}_0) \cdot \left\| \mathbf{f}_\theta(\mathbf{x}_t, t) - \mathbf{v}(\mathbf{x}_0, \mathbf{x}_t, t) \right\|^2 \right]
% \end{equation}

% Hybrid Dataset: The model is trained on a dataset combining synthetic human-product pairs (generated via 3D rendering) and real-world images/videos, augmented with multi-class samples (e.g., 50+ product categories).
% Augmentation Techniques: Data augmentation includes geometric transformations (rotation, scaling), photometric adjustments (brightness, contrast), and category mixing (pairing humans with diverse products).
% Loss Functions: The training objective combines a diffusion loss (L2 loss on denoised features) and an identity preservation loss (perceptual loss for human/product features), ensuring both realism and fidelity.

\section{Experiments}
\label{sec:exp}

\begin{figure*}
  \centering
  \includegraphics[width=0.95\linewidth]{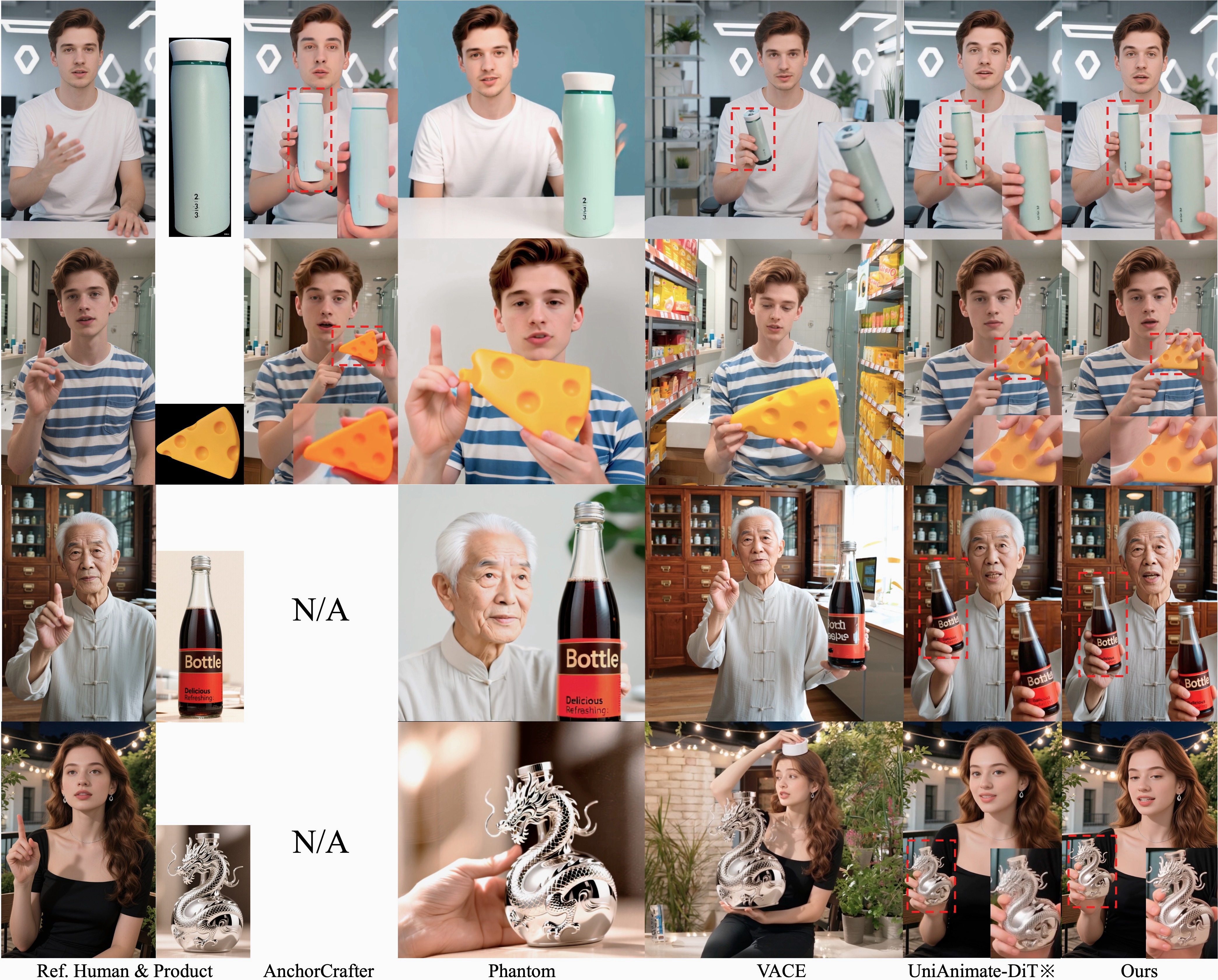} %0.9
  % \vspace{-0.3 cm}
  \caption{Comparisons with AnchorCrafter~\cite{xu2024anchorcrafter}, Phantom~\cite{liu2025phantom}, VACE~\cite{vace} and UniAnimate-DiT※~\cite{wang2024unianimate}. Note that we only generate 3 videos for AnchorCrafter, and UniAnimate-DiT uses our first frames and pose sequences as inputs.}
  \vspace{-0.3 cm}
  % \Description{}
  \label{fig:comparison}
\end{figure*}

\begin{table*}
\centering
\small
\begin{tabular}{lcccccc}
\hline
Methods & CLIP-I$\uparrow$ & DINO-I$\uparrow$ & FaceSim-Arc$\uparrow$ & Motion Smoothness$\uparrow$ & Aesthetic Quality$\uparrow$ & Imaging Quality$\uparrow$ \\
\hline
AnchorCrafter & 0.855 & 0.621 & 0.594 & 0.994 & 0.476 & 0.706 \\
Phantom-WAN2.1-1.3B & 0.875 & 0.795 & 0.640 & 0.992 & 0.493 & 0.730 \\
VACE-WAN2.1-14B & 0.853 & 0.749 & 0.639 & 0.993 & 0.485 & 0.737 \\
UniAnimate-DiT※ & 0.849 & 0.742 & 0.741 & \textbf{0.995} & 0.510 & 0.738 \\
\hline
Ours baseline & 0.826 & 0.731 & 0.759 & 0.993 & \textbf{0.518} & 0.734 \\
Ours w/o text & 0.867 & 0.782 & \textbf{0.779} & 0.994 & 0.513 & 0.736 \\
\textbf{Ours} & \textbf{0.902} & \textbf{0.818} & 0.769 & 0.994 & 0.512 & \textbf{0.739} \\
\hline
\end{tabular}
\caption{Quantitative comparisons with AnchorCrafter~\cite{xu2024anchorcrafter}, Phantom~\cite{liu2025phantom}, VACE~\cite{vace}, UniAnimate-DiT※~\cite{wang2024unianimate}, and our ablation studies (``Ours baseline'' and ``Ours w/o text''). Note that we us our first frames and pose sequences as inputs for UniAnimate-DiT.}
  % \vspace{-0.3 cm}
\label{tab:comparison}
\end{table*}

\begin{table}
\centering
\small
\begin{tabular}{lccc}
\hline
Method & Human & Product & Overall \\
\hline
AnchorCrafter & 12.5 & 77.1 & 58.3 \\
Phantom-WAN2.1-1.3B & 20.8 & 35.7 & 32.1 \\
VACE-WAN2.1-14B & 54.2 & 35.1 & 37.5 \\
UniAnimate-DiT※ & 92.9 & 62.5 & 64.9 \\
\textbf{Ours} & \textbf{100.0} & \textbf{97.0} & \textbf{97.0} \\
\hline
\end{tabular}
\caption{``Good'' or ``same'' rate of various methods in terms of identity preservation for human and product, as well as overall video quality. It should be noted that our first frame and our pose were used as inputs for UniAnimate-DiT.}
  % \vspace{-0.3 cm}
\label{tab:userstudy}
\end{table}

\subsection{Implement Details}

Our training weights are initialized from the pretrained image-to-video Seaweed-7B~\cite{seawead2025seaweed} model, and the training process is conducted using 24 NVIDIA H20 GPUs over two weeks, totaling 100,000 steps. The model operates at a resolution of $720 \times 1280$, with each inference clip containing 65 frames. To ensure full-video consistency, we adopt a sequential generation strategy: the last latent code from the current video segment serves as the initial latent for the subsequent segment. For the appearance guidance module, we set the classifier-free guidance (CFG) parameter to 2.5. Additionally, We use AMO Sampler~\cite{hu2024amo} for the CFG, which can improve the preservation of fine-grained product textures and text details.

\subsection{Comparison}

To mitigate legal risks, all human and product images in this paper, except for three products from AnchorCrafter (we edited their logos), are generated by Seedream 3.0~\cite{gao2025seedream}. To the best of our knowledge, no existing method fully aligns with our input-output pipeline. Thus, we compare our approach with AnchorCrafter~\cite{xu2024anchorcrafter}, which requires multi-view product images and depth maps but is functionally closest, alongside subject-to-video frameworks Phantom~\cite{liu2025phantom} and VACE~\cite{vace}, which better match our input-output design. For evaluating our improvements in dynamic product preservation, we also compare with UniAnimate-DiT~\cite{wang2024unianimate}. As UniAnimate-DiT supports only single human image input with a pose sequence, we only use its self-driving function with our generated first frame and pose sequence. Since it cannot natively generate first frames or perform 3D body retargeting, we denote this variant as UniAnimate-DiT※. Due to AnchorCrafter only releases a trial demo, we can only produce 3 videos using the three objects provided in their demo, and we generated 15 videos for other methods.

In our comparisons, Phantom is trained based on WAN2.1-1.3B~\cite{wan2025}, while both VACE and UniAnimate are based on WAN2.1-14B. According to the report of Seaweed~\cite{seawead2025seaweed}, WAN2.1-14B is a comparable image-to-video generation model with Seaweed-7B adopted in this work. Notably, Phantom and VACE only support landscape video generation, which fails to maintain resolution consistency with other methods. To ensure fairness, we carefully designed our evaluation metrics to minimize the impact of background and human body variations caused by resolution differences among compared methods in subsequent qualitative and quantitative evaluations.

As illustrated in Fig.~\ref{fig:comparison}, our method surpasses others in preserving human and product identities. AnchorCrafter~\cite{xu2024anchorcrafter} demonstrates effective product preservation but requires additional multi-view product images and depth maps. Subject-to-video approaches like Phantom~\cite{liu2025phantom} and VACE~\cite{vace} occasionally fail to respond to prompts (we used the prompt ``This person confidently {holds/graps/lifts} this {product name} in hand, ...''; refer to the supplementary video for more details), and products often appear oversized. By providing UniAnimate with the first frame and retargeted pose, it yields good results but fails to preserve product details, validating the effectiveness of our product preservation module.
As shown in Tab.~\ref{tab:comparison}, we employ CLIP-I~\cite{radford2021learning} and DINO-I~\cite{oquab2023dinov2} to measure product preservation, with our method achieving the best results. Specifically, we detect masks from videos, sample 10 frames per video, crop products using detected masks, and compute cosine similarity with product images via CLIP and DINO. Notably, Phantom and VACE show promising results when ignoring product size. Since full-body representations vary across methods, we use ArcFace~\cite{deng2019arcface} to calculate cosine similarity (for images with a face), where our method also outperforms others, confirming superior human identity preservation. We use VBench~\cite{huang2024vbench} to evaluate ``motion smoothness'', ``aesthetic quality'', and ``imaging quality''. As only the object-holding hand and product move slightly in the videos, all methods exhibit high ``motion smoothness''. Additionally, methods using DiT as a foundation model all perform well in ``aesthetic quality'' and ``imaging quality'', indicating that DiT-based video generation models can generally ensure the video quality.

% We also conduct a user study involving 50 participants using the Good/Same/Bad (GSB) rating scheme (with the exception AnchorCrafter only included 3 videos). During the evaluation, the system randomly presented two videos generated from the same input, and participants were asked to evaluate them on three dimensions: product identity preservation, human identity preservation, and overall video quality. For each dimension, participants selected one of three options: ``good'', ``same'', or ``bad''. Assessments of product and character identity preservation were based on the alignment between reference images and generated video content. The final result was reported as the win rate for each method based on good or same ratings. As shown in Tab.~\ref{tab:userstudy}, our model outperforms all competing methods in all evaluation dimensions.

We also conducted a user study with 50 participants (AnchorCrafter included only 3 videos). The system randomly presented pairs of videos generated from the same input, and participants evaluated them across three dimensions: product identity preservation, human identity preservation, and overall video quality. For each dimension, they chose ``good'', ``same'', or ``bad''. Assessments of product and human identity focused on alignment with reference images. Results were reported as win rates based on ``good'' or ``same'' ratings. As shown in Tab.~\ref{tab:userstudy}, our model outperformed all baselines in all evaluation dimensions.

\begin{figure}
  \centering
  \includegraphics[width=\linewidth]{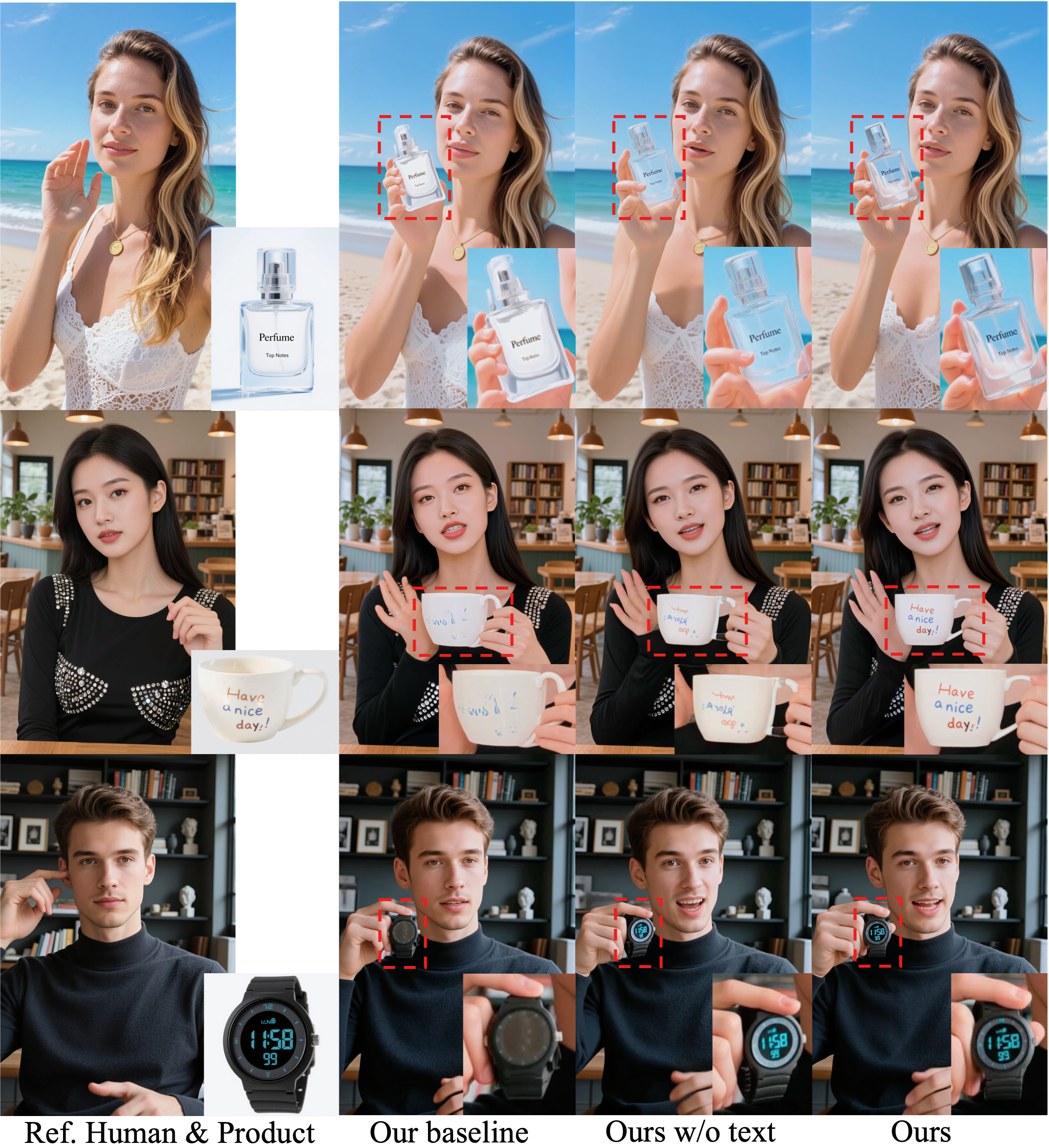} %0.95
  % \vspace{-0.3 cm}
  \caption{Ablation studies with ``Ours baseline'' (w/o object attention and text input) and ``Ours w/o text''.}
  \vspace{-0.3 cm}
  % \Description{}
  \label{fig:ablation}
\end{figure}

\subsection{Ablation Study}
\label{subsec:ablation}
To validate the contributions of our core modules, we conducted ablation experiments comparing three configurations: a baseline trained without object attention or text input (denoted as ``Ours baseline''), a variant trained with object attention but without text input (``Ours w/o text''), and our full model. The results demonstrate that each component plays a critical role in enhancing product preservation and video consistency.

% As shown in Fig.~\ref{fig:ablation}, removing both object attention and text guidance (baseline) leads to significant degradation in product details preservation. Introducing object attention alone improves visual fidelity by directly anchoring product features, but the model still struggles with material properties and geometric consistency, likely due to the lack of semantic context. In contrast, our full model—integrating both object attention and text-derived semantic embeddings—significantly outperforms ``Ours baseline'' and ``Ours w/o text''. The text guidance enables the network to leverage common-sense knowledge about product categories and materials, improving both text preservation and 3D rotation stability. As shown in Tab.~\ref{fig:comparison}, the results also confirm that object attention provides essential visual feature preservation, while text guidance enforces semantic consistency, together enabling robust human-product video generation.

\begin{figure*}
  \centering
  \includegraphics[width=0.95\linewidth]{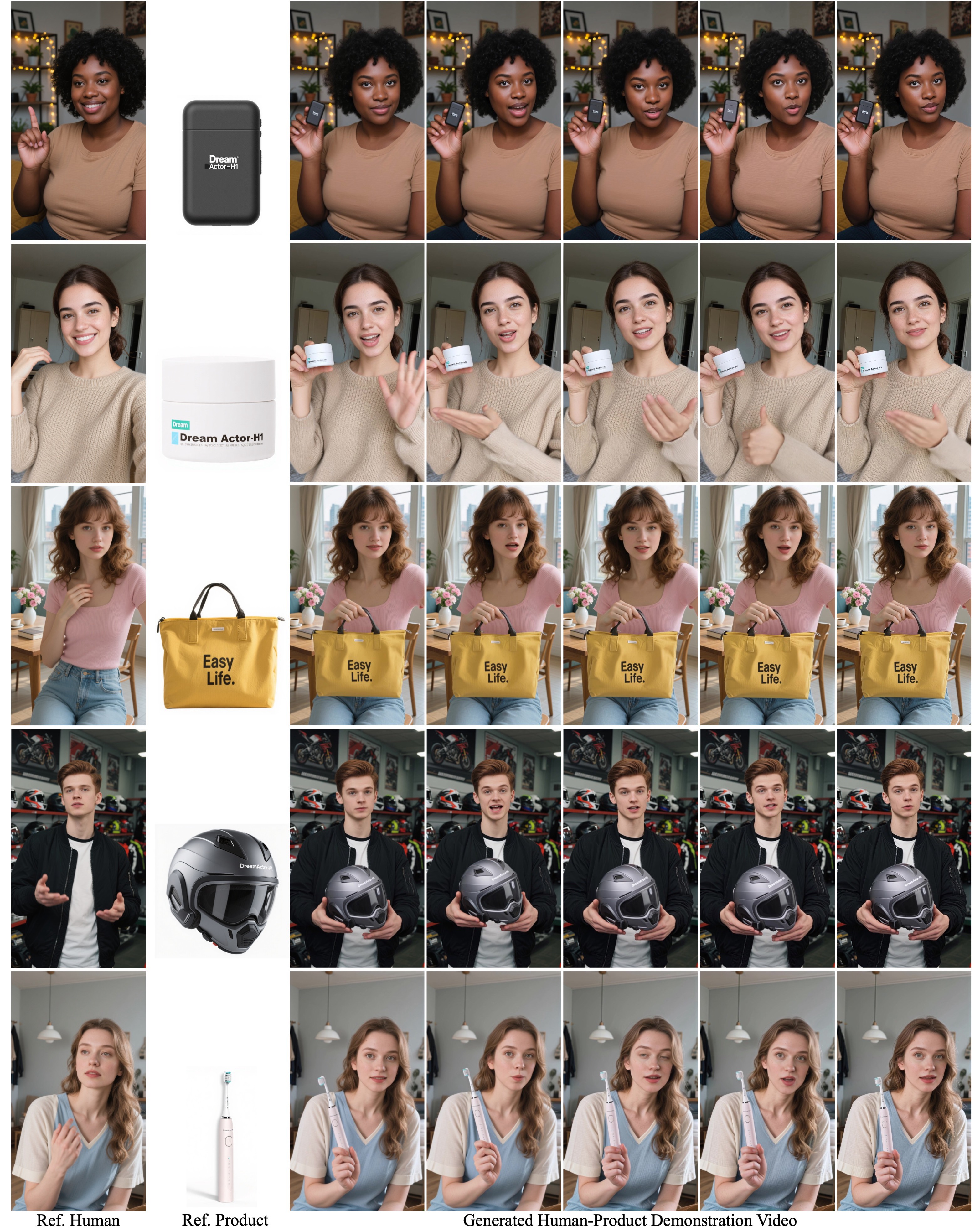}
  \caption{Our video results generated from human images and product images.}
  % \Description{}
  \label{fig-1}
\end{figure*}

\begin{figure*}
  \centering
  \includegraphics[width=0.95\linewidth]{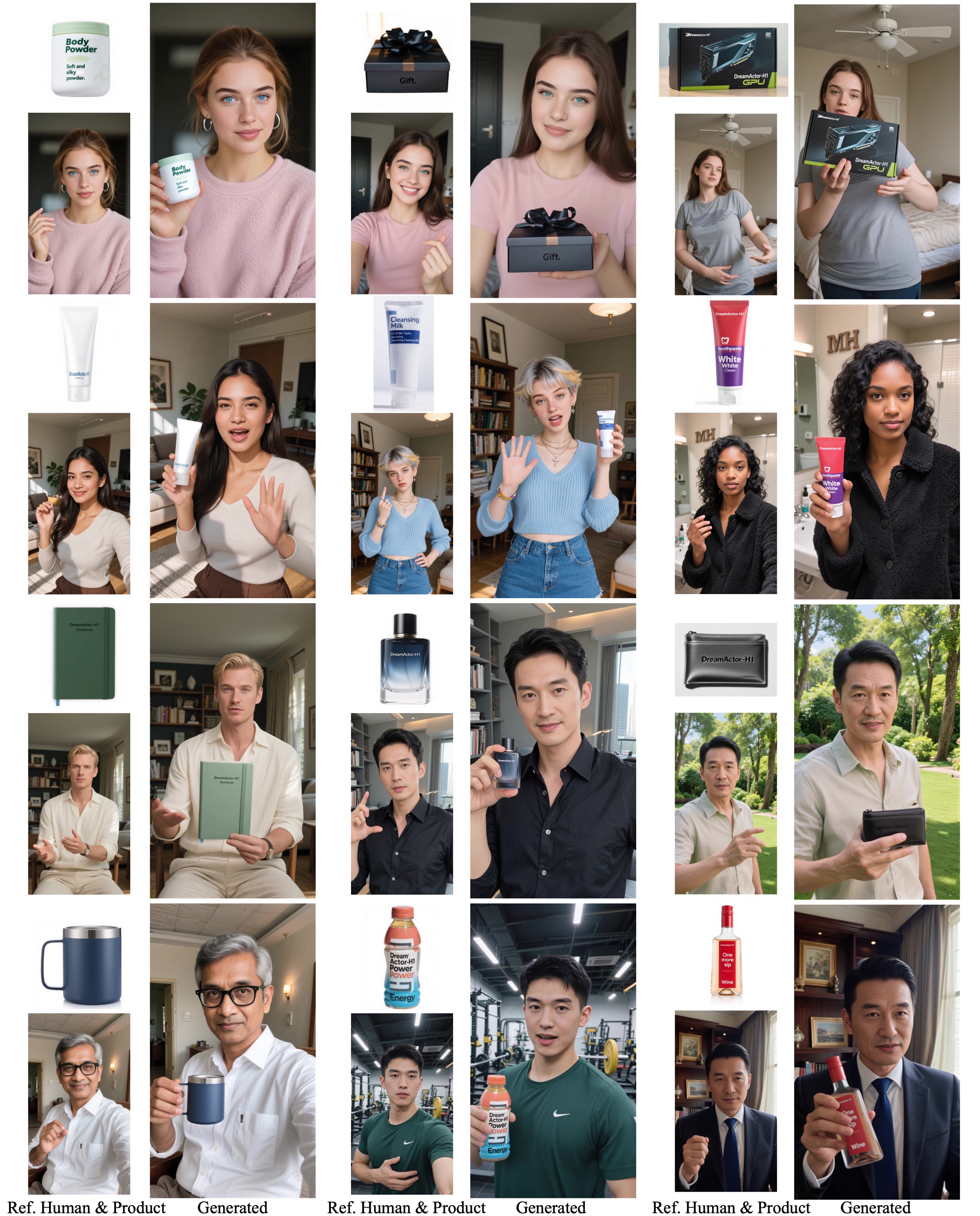}
  \caption{Our results generated from more human images and product images.}
  % \Description{}
  \label{fig-2}
\end{figure*}

As shown in Fig.~\ref{fig:ablation}, removing both object attention and text guidance (baseline) degrades product detail preservation. Introducing object attention alone enhances visual fidelity by anchoring product features, but the model still fails to capture material properties (e.g., transparent surfaces or dial textures), text consistency, and geometric coherence—likely from missing semantic context. In contrast, our full model integrating object attention and text-derived embeddings outperforms both ``Ours baseline'' and ``Ours w/o text''. Text guidance lets the network leverage product-category common-sense to improve text preservation and 3D rotation stability. Tab.~\ref {fig:comparison} also confirms object attention preserves visual features while text guidance ensures semantic consistency, enabling robust human-product video generation. Please refer to Fig.~\ref{fig-1}, Fig.~\ref{fig-2} and our supplementary video for more results.

\section{Discussion}
\label{sec:discussion}

In this work, we present DreamActor-H1, a Diffusion Transformer-based framework that addresses the challenges of generating high-fidelity human-product demonstration videos by integrating masked cross-attention, 3D motion guidance, and semantic-aware text encoding. Our method effectively preserves fine-grained human and product identities while ensuring natural spatial alignment between human gestures and product placements. Extensive experiments demonstrate that DreamActor-H1 outperforms state-of-the-art approaches in maintaining identity integrity and generating physically plausible interactions, making it a solution for e-commerce and digital marketing scenarios.

\noindent\textbf{Limitation.} DreamActor-H1 has several limitations: it currently handles only relatively small-sized products and may produce unnatural interactions due to relying on pre-defined motions unrelated to specific products (e.g. a pose template for picking up products on a table requires that the human reference image includes a table in front of the person, with arms positioned properly relative to the table); the visual-language model (VLM) may inaccurately judge product sizes, especially for non-standard shapes; and it struggles with highly complex product structures, where fine-grained geometry and texture modeling is challenging. Due to the temporal compression of the VAE, text and textures on products may flicker during fast movements. These challenges still need to be addressed in future research.

\noindent\textbf{Ethics considerations.} Human-product image animation has potential social risks, such as misuse for fake product demos or counterfeit promotions. Detection tools~\cite{zhou2024dormant, chen2024demamba} can identify manipulated media. Mitigating risks requires clear ethical guidelines and AI-generated content labeling.
%Human image animation has possible social risks, like being misused to make fake videos. The proposed technology could be used to create fake videos of people, but existing detection tools~\cite{zhou2024dormant,chen2024demamba} can spot these fakes. To reduce these risks, clear ethical rules and responsible usage guidelines are necessary.

\noindent\textbf{Acknowledgement.} We extend our sincere gratitude to Lu Jiang, Yuxuan Luo, Zhengkun Rong, Jiaqi Yang, Tongchun Zuo, Jianwen Jiang, and Youjiang Xu for their invaluable contributions and supports to this research work.

\newpage

{
    \bibliographystyle{ieeenat_fullname}
    \bibliography{reference}
}


\end{document}